\documentclass[conference]{IEEEtran}
\usepackage[utf8]{inputenc}
\usepackage[nottoc]{tocbibind}
\newtheorem{assumption}{Assumption}[section]
\newtheorem{problem}{Problem}[section]
\usepackage[T1]{fontenc}
\usepackage{authblk}
\usepackage[left=0.75in, right=0.75in, top=1in, bottom=0.75in]{geometry}
\usepackage{lipsum}  
\usepackage{amsmath}
\usepackage{amssymb}

\usepackage{graphicx}
\usepackage{subcaption}

\usepackage{cite}
\usepackage{algorithm} 
\usepackage{algpseudocode} 
\usepackage{hyperref}

\title{\LARGE \bf
Deep Learning Based Situation Awareness \\ for Multiple Missiles Evasion
}

\author[1,3]{Edvards Scukins }
\author[2]{Markus Klein}
\author[1]{Lars Kroon}
\author[3]{Petter Ögren}

\affil[1]{\footnotesize Aeronautical Solutions division, SAAB Aeronautics}
\affil[2]{\footnotesize Tactical Control and Data Fusion division, SAAB Aeronautics}
\affil[3]{\footnotesize Robotics, Perception and Learning Lab., Royal Institute of Technology (KTH)}

\begin{document}

%\title{}
\maketitle

\begin{abstract}
As the effective range of air-to-air missiles increases, it becomes harder for human operators to maintain the situational awareness needed to keep a UAV safe.
In this work, we propose a decision support tool to help UAV operators in Beyond Visual Range (BVR) air combat scenarios assess the risks of different options and make decisions based on those. 
Earlier work focused on the threat posed by a single missile, and in this work, we extend the ideas to several missile threats.
The proposed method uses Deep Neural Networks (DNN) to learn from high-fidelity simulations to provide the operator with an outcome estimate for a set of different strategies. Our results demonstrate that the proposed system can manage multiple incoming missiles, evaluate a family of options, and recommend the least risky course of action.

\end{abstract}


%---------------------------------------------------------------------------------------------------------------------------------------------------

\section{Introduction}

Since the First World War, air combat has evolved drastically. Advances in sensors, weapons, and communication allow pilots to engage enemy aircraft at longer and longer distances. Such advances have driven the transition from Within Visual Range (WVR) air combat to Beyond Visual Range (BVR) air combat\cite{stillion2015trends}. In BVR, incoming missiles can fly for up to several minutes, which makes it very difficult for UAV operators to evaluate all incoming data and choose the best course of action. In fact, there is a tendency for the operator to lose track of some of the incoming threats \cite{stillion2015trends}. A support tool that can handle several threats simultaneously and provide an overall analysis is thus needed. Such a tool should support the operators in balancing the risk against the mission goals, as the lowest risk option is often, unfortunately, to ignore the mission entirely, while on the other hand, ignoring the risks ultimately might lead to hefty losses. 

As the flight time of a radar-guided missile can be very long, BVR air combat includes %a planning 
an element that might be compared to real-time strategy games like Starcraft \cite{vinyals2019grandmaster}. %and even Chess or Go \cite{silver2018general}.  
Significant challenges include highly nonlinear dynamics, information uncertainty, and opponents' unknown strategies and goals. 
Onboard sensors can output estimates of opponent locations that depend on the type of hostile aircraft, the electronic warfare countermeasure equipment, and the weather.
%equipment typically delivers measurements of opponent locations as an uncertain estimate. 
However, although precise information is not always available when facing an enemy, operators are typically aware of the capabilities of the enemy's aircraft and weapons systems, and the proposed approach will make use of this information.
\begin{figure}[t!]
    \centering
    \includegraphics[scale=0.6]{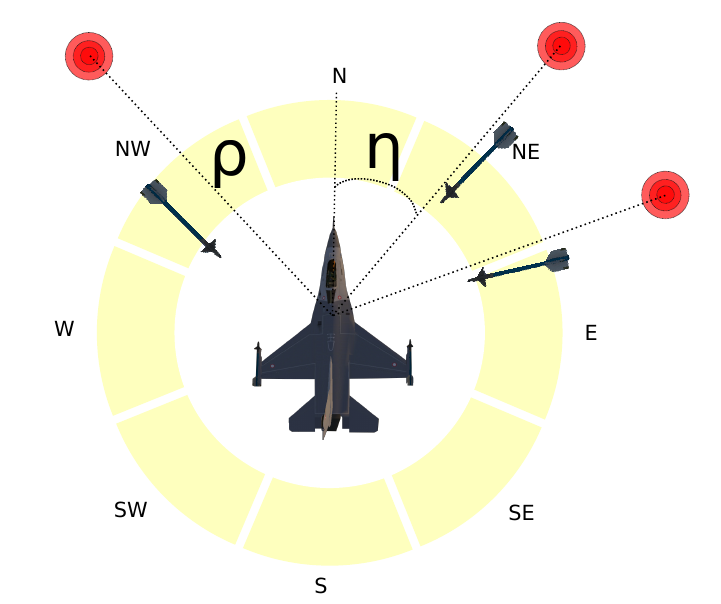}
    \caption{
    Symbolic representation of a situation where the UAV is facing three incoming missiles.
     The exact current locations of the missiles are unknown, but estimates of the time and location of the launches are available. In Figures \ref{fig:testnn_path_Esc3}-\ref{fig:testnn_path_Esc6}, the colored fields around the aircraft icon are used to show the predicted miss distance (MD) of an evasive maneuver in that direction.
     Based on this, the operator can make a trade-off between mission goals and risks when determining what course to choose.
     }
    \label{fig:problem}
\end{figure}

In our previous work \cite{scukins2021using}, we have studied the case of a UAV facing one incoming missile. Using Reinforcement Learning (RL), we computed the optimal evasive maneuver and the resulting miss distance (MD) when performing the maneuver. However, that approach cannot be applied when faced with several adversarial aircraft engaging at the same time. When considering multiple incoming missiles from different angles, the optimal evasive maneuver relative to one missile differs from the other, and you clearly cannot execute two different maneuvers simultaneously.
Furthermore, the most effective evasive action for one pair of incoming missile threats might be determined by solving a specific problem offline and storing the outcome, but this approach becomes impractical due to the vast number of possible threat combinations.

In this paper, we first note that MD estimates are an intuitive risk estimate for human operators. Thus, we would like to present the operators with a set of options such as illustrated in Figure \ref{fig:problem}. The yellow fields in the figure would then be colored according to risk. If executing an evasive maneuver southward gives a MD of 2km it might be colored green, while a westward maneuver might give a miss distance of 0.05km and therefore colored red.

To estimate the MD of a given maneuver in a given direction, when facing multiple threats as described above, we proceed as follows.
First, we learn the MD for a single threat for a set of pre-defined evasive maneuvers in different compass directions. 
Then, as the smallest MD is the one we have to worry about, we simply iterate over all threats, and save the smallest resulting miss distance for each maneuver. 

With this approach, we can provide a decision support tool that provides risk estimates for a family of options without losing track of any detected threats.  Our approach also allows the operator to realize when no safe extraction options are available, such as when threatened from opposite directions at close ranges. Providing decision support for more desperate measures, such as firing all remaining weapons and then losing the UAV, or relying on methods not captured by the model, such as electronic warfare or chaff/flare systems.

The main contribution of this work is thus an approach that enables a UAV operator to assess and handle an arbitrary number of incoming threats, thus extending earlier work considering a single adversarial missile \cite{scukins2021using}. The outline of the work is as follows: A review of the related work is done in Section \ref{sec:RelatedWork}. Section \ref{sec:Background} then provides background on ML and missile guidance, while in Section \ref{sec:ProblemFormulation}, we formally define the problem. The proposed solution is presented in Section \ref{sec:ProposedSolution}, and simulation results are shown in Section \ref{sec:NumericalResults}. Finally, discussion and conclusions are drawn in Section \ref{sec:Conclusion}.

%The main contribution of this work is thus that we provide a deep learning-based risk estimate for a family of evasive options that can handle a scenario with an arbitrary number of incoming threats, thus extending earlier work considering a single adversarial missile \cite{scukins2021using}. The outline of the work is as follows: A review of the related work is done in Section \ref{sec:RelatedWork}. Section \ref{sec:Background} then provides background on ML and missile guidance, while in Section \ref{sec:ProblemFormulation}, we formally define the problem. The proposed solution is presented in Section \ref{sec:ProposedSolution}, and simulation results are shown in Section \ref{sec:NumericalResults}. Finally, discussion and conclusions are drawn in Section \ref{sec:Conclusion}.

%---------------------------------------------------------------------------------------------------------------------------------------------------
%---------------------------------------------------------------------------------------------------------------------------------------------------
\section{Related work}
\label{sec:RelatedWork}

Missile avoidance tactics fall under the category of differential games, in which the state variables of the game change over time. Solving differential games in a decision support context was investigated in \cite{liang2019differential}, where the authors used assumptions on constant speed and simple dynamics to enable the analysis. The authors of \cite{ibragimov2012evasion} studied an evasion game with multiple pursuers and a single evader, although the resources the pursuers use in their pursuit did not exceed the evaders. In \cite{shima2011optimal}, the author studied optimal evasion and pursuit strategies. In this work, the author derived an approach for cooperation between the targeted aircraft, the targeted aircraft's defending missiles, and incoming missiles that are initially targeted toward the aircraft. The main idea is to use the targeted aircraft as a lure to intercept the incoming missile, and perfect information about the state is assumed.

%-> Salvo attack stationary 
In \cite{jeon2010homing}, the authors derived a method for cooperative proportional navigation for a set of missiles to simultaneously reach a designated (stationary) target. 
In 
\cite{hu2019sliding}, the
impact time guiding law design versus non-maneuvering and maneuvering targets tracking time-varying line-of-sight was addressed. The authors of \cite{cho2016modified} derived a closed-form expression for the time-to-go of the pure proportional navigation guidance law, an impact-time-control guidance law, and a cooperative proportional navigation guidance law. In \cite{jiang2023short}, the authors discussed the difficulty of carrying out coordinated air warfare in unmanned aerial vehicle (UAV) swarms. To enhance swarm performance in air combat, the authors developed a multi-agent Transformer introducing virtual objects (MTVO) networks for analyzing air combat circumstances. The authors of \cite{xu2022autonomous} provide a tactical pursuit point (TPP) framework to help tackle the decision-making problem. They combine reinforcement learning with the TPP architecture for a one-to-one dogfight scenario. The approach's primary objective is to get one's aircraft into a better position than the enemy aircraft, i.e., a position that enables using one's own weapons and/or denies the enemy a firing opportunity. In \cite{liles2023improving}, the authors define an Air Battle Management Problem (ABMP) with a friendly aircraft tasked to defend a high-value asset from incoming cruise missiles. The authors formulated the ABMP as a Markov Decision Process (MDP) and utilized approximate dynamic programming (ADP) to generate cost-effective policies for target assignment in the ABMP. %\cite{garcia2021beyond} looked into a BVR scenario in which two competing teams launched an attack on one another's resources. In this paper, the authors devised a zero-sum game to derive vehicle-guiding policies. 
Our work differs from the above in several aspects. 

Firstly, we use a high-fidelity flight dynamics model that is supplied by the JSBSim simulation environment \cite{berndt2004jsbsim} and based on NASA wind tunnel measurements\footnote{\url{https://github.com/JSBSim-Team/jsbsim/tree/master/aircraft/f16}},
whereas many paper listed above rely on highly simplified models of the aircraft.
Simpler models need less computing power, and are easier to investigate analytically.
However, if a risk assessment approach is to be useful in a real UAV system, it needs to be accurate, and then it cannot rely on highly simplified models.
 While authors in \cite{pope2022hierarchical} use JSBSim for WVR air combat, we adapted this simulation environment for a BVR setting. 

Second, all work discussed so far, including \cite{liang2019differential, ibragimov2012evasion, shima2011optimal, jeon2010homing, hu2019sliding, cho2016modified}
assume perfect information about the states of the missiles and the UAV, whereas we only assume estimates are given of the launch locations of the missiles. In a BVR setting, all information regarding adversarial aircraft and missiles will be associated with measurement errors. Additionally, the missile may fly in passive mode, i.e., without emitting any signal, only receiving information about the target's position from the aircraft that fired it. Thus, our approach addresses a problem where the operator does not have knowledge about the missile locations, only an estimate of where the missiles have been fired from. 

Finally, we present a problem formulation that can handle an arbitrary number of incoming threats, thus expanding our earlier work, where the system can only handle a single missile threat \cite{scukins2021using}.

%---------------------------------------------------------------------------------------------------------------------------------------------------
%---------------------------------------------------------------------------------------------------------------------------------------------------
\section{Background}
\label{sec:Background}
This section briefly describes Feed-forward Neural Networks (FNN) and summarizes the missile guidance law Proportional Navigation (PN).
% Machine learning background
\subsection{Feed-forward Neural Networks}
\begin{figure}[h!]
    \centering
    \includegraphics[scale=0.8]{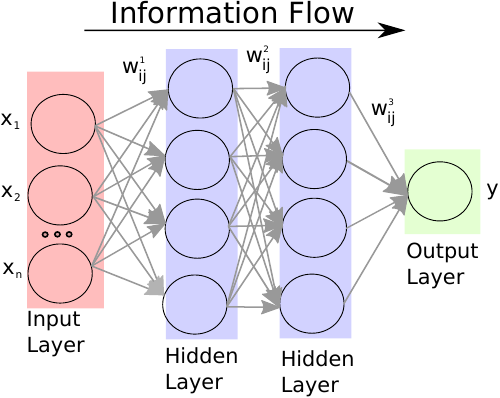}
    \caption{Fully connected Feed-forward Neural Network with n-dimensional input $\Vec{x}$, neural network weights $w^{k}_{ij}$ and output vector $\Vec{y}$.}
    \label{fig:nn}
\end{figure}

FNNs are used for classification tasks and are a subset of Deep Neural Networks (DNNs). In this architecture, the information flows in one direction from the input layer through one or more hidden layers to the output layer, see Figure \ref{fig:nn}. In the field of machine learning (ML), supervised learning is widely used to develop data representations with different levels of abstraction \cite{lecun2015deep}. 
FNN can be represented as a function $f_{FNN}(x; w)$, where $x$ are the input parameters and $w$ represents all learnable parameters within the network. The goal of $f_{FNN}$ is to map a fixed size input $x$ to output $y$ such that $f_{FNN}(x) \approx y$. To do this successfully, an essential task is to gather sizable datasets with associated labels ($\mathcal{L}$). The guidance of the optimization process is done through minimizing the objective function $E = \sum_{n=1}^{N}(y_n-\mathcal{L}_n)^2$, where the network weights are updated iteratively.

In our problem, 
aerodynamic forces acting on the aircraft cause its flight dynamics to be nonlinear. The angle of attack, air density, airspeed, and many other variables may impact the aircraft's dynamics and how various maneuvers are carried out. Due to this, we approximate the nonlinear models using $f_{FNN}$ in conjunction with nonlinear activation functions like $tanh(x)$. The parameters of $w^{k}_{ij}$ are changed using a gradient descent with a learning rate $\alpha$. The weights can be adjusted using the formula $\Delta w^{k}_{ij} = - \alpha \frac{\partial E(x)}{\partial w^{k}_{ij}}$, which takes into account both the target value's error and the FNN's actual output. Additional regularization processes are applied to obtain the classification task's desired performance.

\subsection{Proportional Navigation}

A common control approach for missiles, and other guided weapons, designed to intercept a moving target is the law of proportional navigation (PN). PN has been shown to be effective in various real-world situations, and is based on the idea that the heading of the missile should be changed in proportion to the rate at which the bearing angle changes between it and the target\cite{yanushevsky2018modern}. Thus we have
$a^{PN} = N \dot{\lambda} V$, where $a^{PN}$ is the desired transversal acceleration, $N$ is a positive constant that determines the rate of acceleration,  $\dot{\lambda}$ is the rate of change of the line-of-sight (LOS) angle between the missile and the target, and $V$ is the closing velocity. In this study, we apply a pure PN for missile guidance in combination with increased altitude during mid-course guidance to reduce air drag.

%---------------------------------------------------------------------------------------------------------------------------------------------------
%---------------------------------------------------------------------------------------------------------------------------------------------------
\section{Problem Formulation}
\label{sec:ProblemFormulation}

We denote the adversarial agents as the red team $\mathcal{R}$ of size $N_{\mathcal{R}} \in \mathbb{Z}$ and a single blue agent as $\mathcal{B}$ of size $N_{\mathcal{B}}=1$.
Let the state of the UAV be represented by  $(h, \phi, \theta, \psi, v )\in \mathbb{R}$ and information about the missile launch location be given by $(\rho, \nu, \tau, \eta, \beta )\in \mathbb{R}$, these variables are described in Table~\ref{tab:params}.  

\begin{table}[h] 
\centering 
\begin{tabular}{|c|c|c|c|}
    \hline
    $h$ & UAV altitude  & $\rho$ & Distance to firing position\\
    \hline
    $\phi$ &  UAV roll  & $\nu$  & Launch velocity\\
    \hline
    $\theta$ & UAV pitch& $\tau$ & Time since launch\\
    \hline
    $\psi$ &  UAV heading  & $\eta$ & Relative angle to firing position\\
    \hline
    $v$ & UAV velocity & $\beta$ & Firing altitude\\
    \hline
\end{tabular}
\caption{Description of Variables.} 
\label{tab:params}
\end{table}

%In contrast, the exact location of the $\mathcal{R}$ team is unknown. 
%Suppose the $\mathcal{R}$ team fires missiles toward the $\mathcal{B}$ team. In that case, the approximate firing position/velocity is estimated using a Missile Approach Warning system (MAW) capturing the flash associated with the launch.
%\begin{assumption}
%The agents of the $\mathcal{B}$ can only track the opposing team's location with a significant uncertainty, but the time and position of missile launches can be estimated with a smaller uncertainty. 
%\end{assumption}

%The parameters used to represent the state of the missile launch location are $(\rho, \nu, \tau, \eta, \beta)$. Since we do not have the exact launch location, the observations are represented by a set of normal distributions $\mathcal{N}(\mu,\,\sigma^{2})$, i.e., the mean of the observation and the corresponding variance $((\rho_{\mu}, \rho_{\sigma}), (\nu_{\mu}, \nu_{\sigma}), (\tau_{\mu}, \tau_{\sigma}), (\eta_{\mu}, \eta_{\sigma}), (\beta_{\mu}, \beta_{\sigma}))$.

The main problem addressed in this work is the following: 
\begin{problem}
\label{problem}
In a BVR air combat situation with an arbitrary number of incoming missiles and a family of evasive maneuvers, what is the estimated miss distance (MD) associated with each maneuver, and which is the safest from a given set of maneuvers?
\end{problem}

Since air combat, in general, contains a large number of complex components,   we make an additional assumption regarding the problem setup.

\begin{assumption}
We have good estimates of the missile type, including dynamics and guidance algorithms, used by the opposing team. 
\end{assumption}
This is a reasonable assumption since significant effort is often spent to analyze different missiles, and most of them apply some form of PN guidance law with slightly higher altitudes in mid-phase to increase range.
However, if the assumption does not hold and adversarial missiles are better than estimated, the predicted miss distances will be too optimistic.
\begin{assumption}
The time that the missile can remain in the air, $t_{f}^{M}$, before hitting the ground due to fuel shortage, is smaller than the time the UAV can remain in the air, $t_{f}^{UAV}$.
%We let $t$ be the time variable. We assume that a missile is a system with a finite time horizon, from $t_0$ to $t_f$, where $t_0$ and $t_f$ represent initial and terminal time, respectively. Given that aircraft, in general, have a longer flight time ($t_{f}^{UAV}$) than an air-to-air missile flight time ($t_{f}^{M}$), the following is assumed $t_{f}^{M} < t_{f}^{UAV}$. 
\end{assumption}

%---------------------------------------------------------------------------------------------------------------------------------------------------
%---------------------------------------------------------------------------------------------------------------------------------------------------
\section{Proposed Solution}
\label{sec:ProposedSolution}
To solve Problem \ref{problem}, we collect data from executing a family of single-missile evasive maneuvers moving in different compass directions, then use DL to train a model to predict the MD for a single missile, and finally use this model to evaluate each maneuver against a set of threats. The UAV operator can then check which options are sufficiently safe and consider his mission goals when choosing the best option.

\subsubsection{Data Collection and Training}
We let the set of evasive maneuvers be described by 
$\{ \pi_{1}, \pi_{2} ... \pi_{n}  \} = \Pi$, where each policy denotes an evasive maneuver in a specific direction. 
To collect data,  we create an environment $E$, where the UAV faces an incoming missile fired by a single aircraft from the $\mathcal{R}$ team. For each policy $\{ \pi_{1}, \pi_{2} ... \pi_{n}  \}$, we collect the corresponding states from the environment: $s_t$ = $(h, \phi, \theta, \psi, v, \rho, \nu, \tau, \eta, \beta)$. 
\begin{algorithm}
\caption{Data Collection: Environment $E$, Policy Group $\Pi$, Number of episodes $N$, Data buffer $D_{\Pi}$}\label{pse:data_collect}
\begin{algorithmic}
\For $ \quad \pi_i\leftarrow 1$ ...  $|\Pi|$ \Comment{Initialize policy $\pi_i$}
\For $ \quad s_{o \: i}\leftarrow 1$ ...  $N$   \Comment{Initialize initial state $s_0$}
\State Initialize $E(s_{o \: i})$ \Comment{Reset Environment}
\While{State Not Terminal} \Comment{Run simulation}
\If{$s_{t}$ is terminal state $(s_{T})$ } 
    \State Add $s_{T}$ tuple $D_{\pi_i}$\Comment{Add label to $s_{0},..,s_{T}$}
\ElsIf{$s_{t}$ is not terminal state}
    \State Add tuple $(s_t)$ to $D_{\pi_i}$ \Comment{Append state}
\EndIf
\EndWhile
\EndFor
\State Add $D_{\pi_i}$ to $D_{\Pi_i}$ \Comment{Append $D_{\pi_i}$ to policy family $D_{\Pi_i}$}
\EndFor
\end{algorithmic}
\end{algorithm}
The outcome in terms of MD (the closest distance between the UAV and the missile) of following each policy $\{ \pi_{1}, \pi_{2} ... \pi_{n}  \}$ is later used as a label. $ MD=0$ means that the missile hit the target, and $MD > 0$ means that the missile missed the target. The data collection procedure is summarized in Algorithm \ref{pse:data_collect}. Note that we assume exact knowledge of the parameters used for state representation during the data collection process. 
%, but when using the tool, we can account for uncertainties by taking several samples from the corresponding distributions.

Algorithm \ref{pse:training} describes the training procedure to obtain a system that estimates the threat of a single missile in a given situation.
\begin{algorithm}
\caption{$f_{FNN}$ Training: Data set $D_{\Pi}$, Number of Epochs $Eps$, Set of $f_{FNN}$ models $M_{\Pi}$}\label{pse:training}
\begin{algorithmic}
\For $ \quad D_{\pi i} \leftarrow D_{\Pi}$ \Comment{Go through all data in $D_{\Pi}$}
\State $x,y \gets$ $D_{\pi i}$ \Comment{Extract states and labels}
\State $f_{FNN}^{\pi_i} \gets$ $M_{\Pi}$ \Comment{Initialize $f_{FNN}^{\pi_i}$}
\State $Optimizer, Loss \gets$ Initialize optimizer and loss
\For $\quad Eps$ \Comment{Iterate through the data set}
\State Reset gradients of $Optimizer$
\State Get $\hat{y}\gets f_{FNN}^{\pi_i} (x)$ \Comment{Get estimate}
\State Get $Loss$ \Comment{Calculate loss between $\hat{y}$ and $y$}
\State Backpropagate $Loss$ \Comment{Compute gradients}
\State Update model parameters using $Optimizer$ 
\EndFor
\EndFor
\end{algorithmic}
\end{algorithm}

Executing both Algorithms \ref{pse:data_collect}, \ref{pse:training}, we obtain an $f_{FNN}^{\pi_i}$ model to estimate an approximate outcome when the UAV faces a single incoming missile, executing a given maneuver. Each $f_{FNN}^{\pi_i} (s_t)$ can estimate an outcome from a given state $s_t$ when following policy $\pi_i$. The next part describes how we can use the $f_{FNN}^{\pi_i}$ to estimate the risk of N different incoming missiles.

\subsubsection{Deep Learning Based Situation Awareness}
 When faced with many dangers, it is common for operators to concentrate on the most serious one, thereby often overlooking other threats, which might also be serious. To avoid such issues
we apply Algorithm \ref{pse:sol}, where we first obtain our state with respect to all fired missiles, $S_E$. Then, we query the $f_{FNN}^{\pi_i} (s_t)$ to see the smallest expected MD. For every policy $\pi_i$, there is a set of MDs (given multiple threats); then, we select the smallest one to represent the MD value of that policy. Finally, if we wish to find the safest evasive policy, we select the policy with the largest value out of the remaining MDs.

\begin{algorithm}
\caption{Situation Awareness: State of the environment $S_E$, Set of $f_{FNN}^{\pi_i}$ models $M_{\Pi}$}\label{pse:sol}
\begin{algorithmic}
\State $M_{\Pi} \leftarrow$ Initialize with pre-trained parameters 
\State $S_E \leftarrow$ Get all missile launches locations 
\State $\mathcal{O}_{\pi i}$ $\leftarrow$ Initialize tuple for MD \Comment{ MD Buffer}
\State $\mathcal{O}$ $\leftarrow$ Init. tuple for SA \Comment{ Situation Awareness Buffer}
\For $\quad s_i \leftarrow 1$ ...  $S_E$    \Comment{Go through each firing location}
\For $f_{FNN}^{\pi_i}$ $\leftarrow$ 1 ...  $M_{\Pi}$   \Comment{Iterate policies}
\State $ y \gets$ $f_{FNN}^{\pi_i}(s_i)$ \Comment{Estimate MD for policy $\pi_i$}
\State Add $y$ to tuple $\mathcal{O}_{\pi i}$  \Comment{Collect MD for policy $\pi_i$}
\EndFor
\State Add $\mathcal{O}_{\pi i}$ to tuple $\mathcal{O}$  \Comment{Collect all estimated MD}
\EndFor
\For $\quad \mathcal{O}_{\pi i} \leftarrow 1$ ...  $\mathcal{O}$   \Comment{Go through each estimate}
\State $\mathcal{O}_{\pi i}^{min} \leftarrow min(\mathcal{O}_{\pi i})$ \Comment{Lowest MD for policy $\pi_i$}
\State Update SA with $\mathcal{O}_{\pi i}^{min}$ \Comment{Present to UAV operator}
\EndFor
\State $\pi^{*}$ $\leftarrow$ $max(\mathcal{O}_{\pi 1}^{min}, \mathcal{O}_{\pi 2}^{min}....\mathcal{O}_{\pi n}^{min})$\Comment{Safest evasive policy with largest MD}
\end{algorithmic}
\end{algorithm}

\subsubsection{Presenting the estimated MDs to the UAV operator}

We want the UAV operator to have an instant SA of the incoming threats and how they relate to his mission goals.
Therefore, we propose a circular diagram as shown in Figure \ref{fig:testnn_path_Esc3} with an aircraft icon in the center indicating the present heading. The color of each heading illustrates the predicted MD of an evasive maneuver in that direction.
Based on the mission goals, the operator might very well fly in the direction of increased threat for some time, but when the evasive maneuvers in that direction turn yellow or red, it might be good to start heading in the direction of the safer options or even execute the safest evasive maneuver.

\section{Numerical Results}
\label{sec:NumericalResults}

First, we discuss the method and the parameters used to train the FNN models. Then, we look at three scenarios where an operator must evade three, four, and six incoming missiles, see Figures \ref{fig:testnn_path_Esc3}, \ref{fig:testnn_path_NoEsc}, and \ref{fig:testnn_path_Esc6}. These examples illustrate how an arbitrary number of threats can be handled, and how the SA is presented to the operator.

\subsubsection{Training the Feed-forward Neural Network} We begin by creating a collection of pre-defined policies $\{ \pi_{N}, \pi_{NE}, \pi_{E}, \pi_{SE}, \pi_{S}, \pi_{SW}, \pi_{W}, \pi_{NW} \} = \Pi$, where each policy denotes an evasive maneuver in a specific compass direction. 
Here we note that even though all presented figures might look planar, the simulations are in full 3D, and the evasive maneuvers all involve descending in altitude, as it increases speed and also gives the incoming missile increased drag.
However, we note that the method can be used with any family of policies $\Pi$, deemed relevant by the user.

We initiate data collection by simulating an unmanned version of the F16 (team $\mathcal{B}$) aircraft facing a single incoming missile (M1) from various initial states fired by an aircraft from team $\mathcal{R}$. The variations of the initial conditions are summarized in Table \ref{tab:init_val}. 
\begin{table}[b!]
\centering
\caption{Table of initial conditions.}
\begin{tabular}{p{5.5cm} l   r }
\hline
Parameter & Value \\ 
\hline
Initial Velocity: F16$\mathcal{B}$  & $300 - 365 $ [m/s]  \\
Initial Altitude: F16$\mathcal{B}$  & $6000 - 10000 $ [m]  \\
Initial heading: F16$\mathcal{B}$  & $0-360 $ [deg] \\
Initial pitch/roll: F16$\mathcal{B}$  & $0 , 0$ [deg] \\
Initial Velocity: M1$\mathcal{R}$   & $280 - 320 $ [m/s]  \\ 
Initial Altitude: M1$\mathcal{R}$   & $9000 - 11000 $ [m]  \\ 
Firing Distance: M1$\mathcal{R}$    & $40 - 80$ [km] \\ 
\hline
\end{tabular}
\label{tab:init_val}
\end{table}
At each time-step, we collect the state $s_t$ = $(h, \phi, \theta, \psi, v, \rho, \nu, \tau, \eta, \beta)$ with a perfect knowledge for all the state parameters. Each scenario ends when the missile hits the target, misses it, or when the missile loses enough speed to make interception impossible. If the missile intercepts the target, an MD of $0$ is recorded for all encountered states during the evasive maneuver. If the missile becomes inactive, i.e., can no longer intercept the target, the closest encountered distance is given as MD.  
For each policy within $\Pi$, we simulate 50,000 episodes, resulting in data containing roughly 5.5 million states for each policy. The FNN architecture used in this study is summarized in Table \ref{tab:network_architecture}. The training process used 400 Epochs, a learning rate of $0.0003$, a batch size of 64, and an MSE loss function. 
\begin{table}[h!]
\centering
\begin{tabular}{c c c c }
\hline
Layer & Input Size & Output Size & Activation Function \\
\hline
FC1 & 10 & 128 & tanh \\
FC2 & 128 & 256 & tanh \\
FC3 & 256 & 256 & tanh \\
FC4 & 256 & 64 & tanh \\
FC5 & 64 & 1 & None \\
\hline
\end{tabular}
\caption{Feed Forward Neural network architecture.  FC1, FC2, $\ldots$ FC5 are all fully connected layers, and each is followed by a tanh activation function, except for the last one, which has no activation function.}
\label{tab:network_architecture}
\end{table}

%--------------------------------------------------------------------------------------------------------------------------------------
\subsubsection{Three missile scenario}

\begin{figure}[h!]
    \centering
    \includegraphics[scale=0.60]{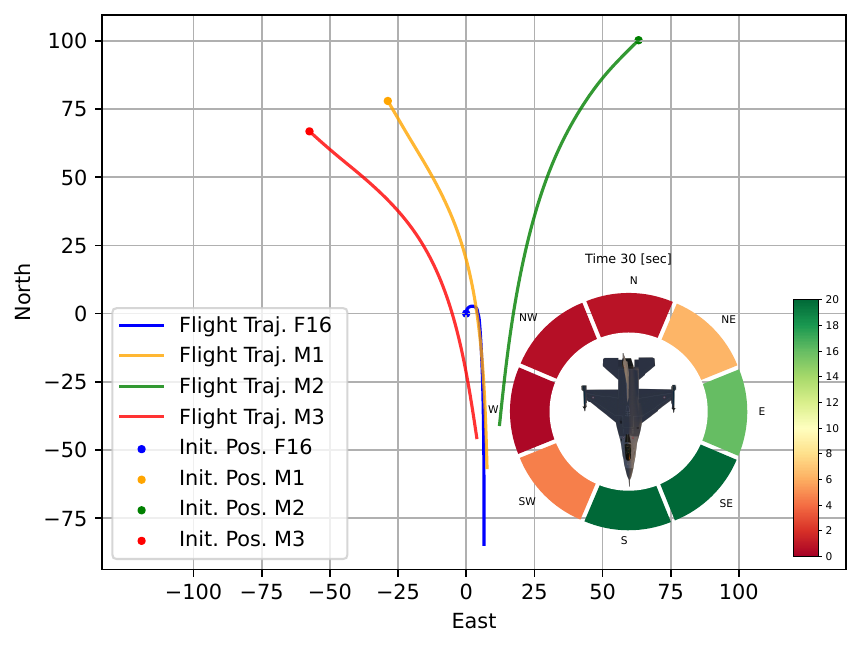}
    \caption{The unmanned F16 faces three incoming missiles, with the starting location indicated by dots. As can be seen, the south-eastern options are safe (green circle segments), while the north-western ones are not (red circle segments). If the mission objectives required a northeastern course, that would have been possible, but with a considerably smaller miss distance (orange circle segments). In this simulation, the operator chose the safest maneuver $\pi_S$, flying south. }
    \label{fig:testnn_path_Esc3}
\end{figure}

We consider a situation where there are three F16$\mathcal{R}$ aircraft closing in on a single unmanned F16$\mathcal{B}$ aircraft, see Figure \ref{fig:testnn_path_Esc3}. We presume that the F16$\mathcal{B}$ aircraft has been tracking the enemy planes before their missile launch and that onboard sensors like the Missile Approach Warning System (MAW) detect a flash from the missile ignition. Such systems may have difficulty determining the precise location of the missiles but can approximate the launch position. Thus, the state $s_t$ only includes variables that might be accessible for assessment during air combat.

After the missile launch, we execute Algorithm \ref{pse:sol}. We query each $f_{FNN}^{\pi_i}(s_i)$ with respect to each missile to obtain MD estimates. Given three missiles, we obtained three estimates from which we chose the lowest $\min(f_{FNN}^{\pi_i} (s_1), f_{FNN}^{\pi_i} (s_2), f_{FNN}^{\pi_i} (s_3)):=  \mathcal{O}_{\pi i}^{min}$ to represent the expected outcome by following policy $\pi_i$. Figure \ref{fig:testnn_path_Esc3} illustrates an evasive maneuver following the highest expected MD for all policies $\pi^{*}$ $\leftarrow$ $max(\mathcal{O}_{\pi N}^{min}, \mathcal{O}_{\pi NE}^{min}....\mathcal{O}_{\pi NW}^{min})$. In this three-missile case, the $\pi_{S}$ is the safest evasive policy that maximizes MD. 

Note that it might not always be desirable to execute the safest policy.
Looking at the SA of the colored ring in Figure \ref{fig:testnn_path_Esc3}, we see that it is also quite safe to fly eastwards.
From a mission goal point of view, this might be better, since it might keep one or more of the hostile aircraft in the field of view of the nose-mounted radar.

\begin{figure}
    \centering
    \includegraphics[scale=0.6]{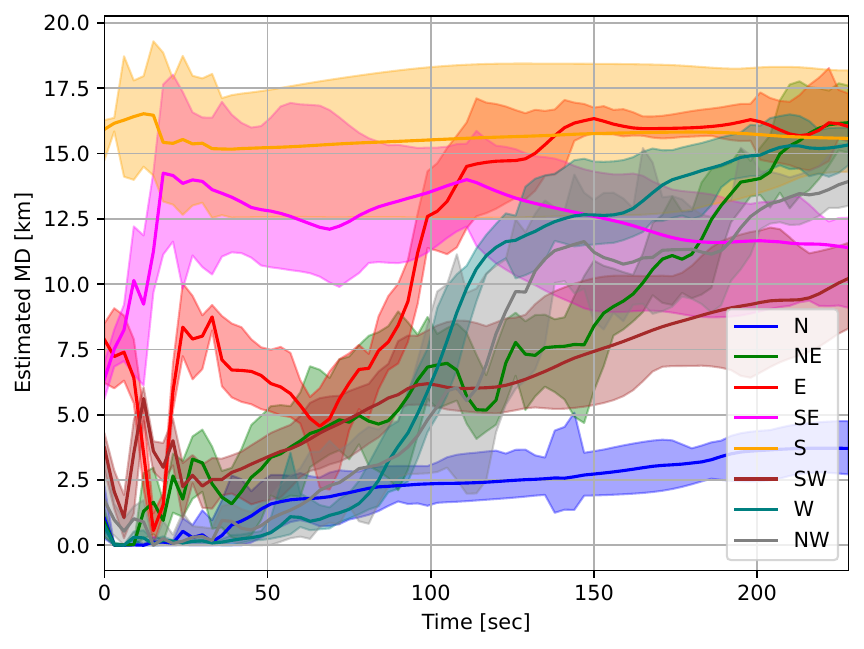}
    \caption{An illustration of the estimated MD distributions, given the sensor variances of Table \ref{tab:var_obs}, over time for each flight direction from the scenario in Figure \ref{fig:testnn_path_Esc3}. The plotted MD circles in the other figures can be seen as snapshots from a given time instant of this plot. Note that the operator applies $\pi_S$ and thus the estimates for that maneuver remains roughly constant over time. The operator can continuously balance these risk estimates with other mission objectives to decide the proper course of action.}
    \label{fig:testnn_est_fill}
\end{figure}

In air combat,  onboard sensors only provide estimates of data such as target locations.  
Assuming that the magnitudes of the uncertainties are as listed in Table \ref{tab:var_obs}, we can draw samples from the distributions and run them through our FNNs.
The resulting distributions of the MDs are shown in Figure \ref{fig:testnn_est_fill}. Now, one can either use the averages to draw the colored rings or pick some desired degree of certainty, giving a conservative  SA support of the form: With 90\% certainty, the MD will be larger than 5km.

\begin{table}[h!]
\centering
\caption{Table of the variance for corresponding observations.}
\begin{tabular}{p{5.5cm} l   r }
\hline
Parameter & Value \\ 
\hline
Distance variance $\sigma_{\rho}$   & $100 $ [m]  \\
Launch Time variance $\sigma_{\tau}$   & $1 $ [sec]  \\
Relative angle variance $\sigma_{\eta}$   & $0.1 $ [deg]  \\
Altitude variance $\sigma_{\beta}$   & $100 $ [m]  \\
Initial velocity variance $\sigma_{\nu}$   & $5$ [m/s]  \\
\hline
\end{tabular}
\label{tab:var_obs}
\end{table}

\subsubsection{Four and six missiles scenario}

\begin{figure}[h!]
    \centering
    \includegraphics[scale=0.6]{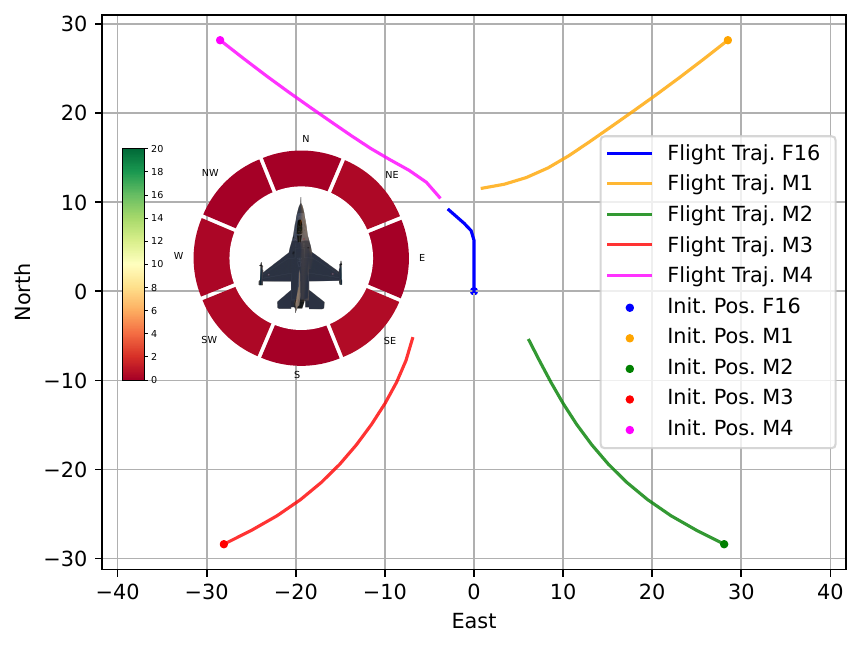}
    \caption{The unmanned F16$\mathcal{B}$ faces four incoming missiles fired at close range, with dots indicating the starting positions. 
    All policies consistently predict a hit at this range (the circle is red), so there are no safe evasive maneuvers.
    In such a case, the operator might fire all remaining weapons in a final effort or hope that electronic warfare options such as chaffs or flares might save the UAV in the last second.
    %Following the safest policy ($\pi_N$, $\pi_NW$) results in a missile hit after 35 seconds, as predicted by the estimated MD.
    }
    \label{fig:testnn_path_NoEsc}
\end{figure}

Consider four F16$\mathcal{R}$ aircraft approaching  from the North-East, South-East, South-West, and North-West, and the unmanned F16$\mathcal{B}$  in the middle, flying Northbound, as illustrated in Figure \ref{fig:testnn_path_NoEsc}. By executing Algorithm \ref{pse:sol}, we obtain an SA represented by the circular diagram in Figure \ref{fig:testnn_path_NoEsc}. Since the missiles were fired from a significantly shorter range, no policy within $\Pi$ produces a non-zero MD.  
Finally, 34 seconds after launch, the M4 missile hits the UAV. 

\begin{figure}[h!]
    \centering
    \includegraphics[scale=0.6]{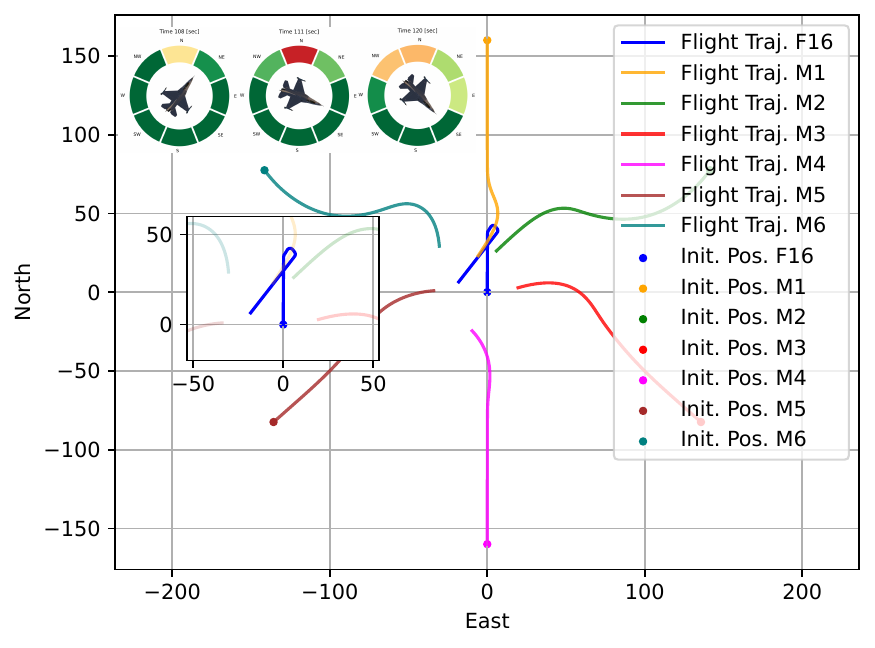}
    \caption{The unmanned F16 faces six incoming missiles, fired at very long range. As can be seen, after flying north initially, only the northern option is dangerous. Thus the operator turns south and then southwest to keep the UAV safe.}
    \label{fig:testnn_path_Esc6}
\end{figure}

Consider now six F16$\mathcal{R}$ aircraft distributed around a single F16$\mathcal{B}$ aircraft, see Figure \ref{fig:testnn_path_Esc6}. In this scenario, we have placed the $\mathcal{R}$ team out of range, i.e.,  the missiles can only reach the target if the target decreases the distance by flying towards the launched missile. Having this information, an intuitive strategy would be to circle in place, assuming the adversaries will not close in for an additional shot. 
Since there was no data with such a large distance (only distances between 40-80 km) during the training process, the FNN provided us with an overconfident estimate. When flying North for 50 km, we notice that other policies have become more favorable compared to the Northern direction. In this situation, following the safest evasive path $\pi^{*}$ $\leftarrow$ $max(\mathcal{O}_{\pi N}^{min}, \mathcal{O}_{\pi NE}^{min}....\mathcal{O}_{\pi NW}^{min})$ the aircraft manages to evade all the incoming missiles.

\section{Discussion}
\label{sec:Discussion}

Based on interviews conducted with US Air Force pilots \cite{endsley1993survey}, a good SA should inform the pilot of the \emph{who, when, and where} of allies and dangers in a given scenario. The proposed method is a step in this direction, as it provides an instant picture of the current threats in a human-readable form, that allows the operator to make tradeoffs between mission goals and risk exposure of the UAV. 

The approach is also executable in real-time on modest hardware, as it takes 
0.178 seconds to generate the SA estimate for the six missile cases above, on an Intel Core i7-8700CPU@3.20GHz x 12 processors.

%----------------------------------------------------------------------
\section{Conclusion}
\label{sec:Conclusion}

In this paper, we propose an approach to estimate the risk of a UAV facing several incoming missiles and illustrate it with three different examples.
Such a risk estimate is important to a UAV operator, trying to balance risk exposure with mission objectives.

The approach goes beyond earlier work by handling multiple threats, using high-fidelity flight simulator dynamics, not assuming full knowledge of the states of the threats, and handling uncertainties in the estimates of the launch data that is used.

\section*{Acknowledgment}
The authors gratefully acknowledge funding from Vinnova, NFFP7, dnr 2017-04875.

\bibliographystyle{unsrt}
\bibliography{bibliography}

\begin{thebibliography}{10}

\bibitem{stillion2015trends}
John Stillion.
\newblock {\em Trends in air-to-air combat: Implications for future air
  superiority}.
\newblock Center for Strategic and Budgetary Assessments., 2015.

\bibitem{vinyals2019grandmaster}
Oriol Vinyals, Igor Babuschkin, Wojciech~M Czarnecki, Micha{\"e}l Mathieu,
  Andrew Dudzik, Junyoung Chung, David~H Choi, Richard Powell, Timo Ewalds,
  Petko Georgiev, et~al.
\newblock Grandmaster level in starcraft ii using multi-agent reinforcement
  learning.
\newblock {\em Nature}, 575(7782):350--354, 2019.

\bibitem{scukins2021using}
Edvards Scukins and Petter {\"O}gren.
\newblock Using reinforcement learning to create control barrier functions for
  explicit risk mitigation in adversarial environments.
\newblock In {\em 2021 IEEE International Conference on Robotics and Automation
  (ICRA)}, pages 10734--10740. IEEE, 2021.

\bibitem{liang2019differential}
Li~Liang, Fang Deng, Zhihong Peng, Xinxing Li, and Wenzhong Zha.
\newblock A differential game for cooperative target defense.
\newblock {\em Automatica}, 102:58--71, 2019.

\bibitem{ibragimov2012evasion}
Gafurjan~I Ibragimov, Mehdi Salimi, and Massoud Amini.
\newblock Evasion from many pursuers in simple motion differential game with
  integral constraints.
\newblock {\em European Journal of Operational Research}, 218(2):505--511,
  2012.

\bibitem{shima2011optimal}
Tal Shima.
\newblock Optimal cooperative pursuit and evasion strategies against a homing
  missile.
\newblock {\em Journal of Guidance, Control, and Dynamics}, 34(2):414--425,
  2011.

\bibitem{jeon2010homing}
In-Soo Jeon, Jin-Ik Lee, and Min-Jea Tahk.
\newblock Homing guidance law for cooperative attack of multiple missiles.
\newblock {\em Journal of guidance, control, and dynamics}, 33(1):275--280,
  2010.

\bibitem{hu2019sliding}
Qinglei Hu, Tuo Han, and Ming Xin.
\newblock Sliding-mode impact time guidance law design for various target
  motions.
\newblock {\em Journal of Guidance, Control, and Dynamics}, 42(1):136--148,
  2019.

\bibitem{cho2016modified}
Namhoon Cho and Youdan Kim.
\newblock Modified pure proportional navigation guidance law for impact time
  control.
\newblock {\em Journal of Guidance, Control, and Dynamics}, 39(4):852--872,
  2016.

\bibitem{jiang2023short}
Feilong Jiang, Minqiang Xu, Yuqing Li, Hutao Cui, and Rixin Wang.
\newblock Short-range air combat maneuver decision of uav swarm based on
  multi-agent transformer introducing virtual objects.
\newblock {\em Engineering Applications of Artificial Intelligence},
  123:106358, 2023.

\bibitem{xu2022autonomous}
Jiwen Xu, Jing Zhang, Lingyu Yang, and Chang Liu.
\newblock Autonomous decision-making for dogfights based on a tactical pursuit
  point approach.
\newblock {\em Aerospace Science and Technology}, 129:107857, 2022.

\bibitem{liles2023improving}
Joseph~M Liles~IV, Matthew~J Robbins, and Brian~J Lunday.
\newblock Improving defensive air battle management by solving a stochastic
  dynamic assignment problem via approximate dynamic programming.
\newblock {\em European Journal of Operational Research}, 305(3):1435--1449,
  2023.

\bibitem{berndt2004jsbsim}
Jon Berndt.
\newblock Jsbsim: An open source flight dynamics model in c++.
\newblock In {\em AIAA Modeling and Simulation Technologies Conference and
  Exhibit}, page 4923, 2004.

\bibitem{pope2022hierarchical}
Adrian~P Pope, Jaime~S Ide, Daria Mi{\'c}ovi{\'c}, Henry Diaz, Jason~C Twedt,
  Kevin Alcedo, Thayne~T Walker, David Rosenbluth, Lee Ritholtz, and Daniel
  Javorsek.
\newblock Hierarchical reinforcement learning for air combat at darpa's
  alphadogfight trials.
\newblock {\em IEEE Transactions on Artificial Intelligence}, 2022.

\bibitem{lecun2015deep}
Yann LeCun, Yoshua Bengio, and Geoffrey Hinton.
\newblock Deep learning.
\newblock {\em nature}, 521(7553):436--444, 2015.

\bibitem{yanushevsky2018modern}
Rafael~T Yanushevsky.
\newblock {\em Modern missile guidance}.
\newblock CRC Press, 2018.

\bibitem{endsley1993survey}
Mica~R Endsley.
\newblock A survey of situation awareness requirements in air-to-air combat
  fighters.
\newblock {\em The International Journal of Aviation Psychology},
  3(2):157--168, 1993.

\end{thebibliography}

\end{document}